\pdfoutput=1
\documentclass[10pt,twocolumn,letterpaper]{article}

\usepackage[margin=0.75in]{geometry}
\usepackage[utf8]{inputenc}
\usepackage[T1]{fontenc}
\usepackage[english]{babel}
\usepackage{times}
\usepackage{microtype}
\usepackage{url}

\usepackage{amsmath}
\usepackage{amssymb}
\usepackage{amsfonts}
\usepackage{mathtools}
\usepackage{amsthm}
\usepackage{graphicx}
\usepackage{booktabs}
\usepackage{enumitem}
\usepackage{tabularx}
\usepackage{multirow}
\usepackage{xcolor}
\usepackage{soul}
\usepackage{tikz}
\usepackage{tcolorbox}
\usepackage[numbers,sort&compress]{natbib}
\usepackage[colorlinks=true,linkcolor=blue!82!black,citecolor=black,urlcolor=blue!73!black]{hyperref}

\tcbuselibrary{skins,breakable}
\usetikzlibrary{shapes.geometric,arrows.meta,positioning,calc}

\DeclareUnicodeCharacter{00E9}{\'e}
\DeclareUnicodeCharacter{00F6}{\"o}
\DeclareUnicodeCharacter{0151}{\H{o}}
\DeclareUnicodeCharacter{2013}{--}
\DeclareUnicodeCharacter{2014}{---}
\DeclareUnicodeCharacter{2019}{'}
\DeclareUnicodeCharacter{201C}{``}
\DeclareUnicodeCharacter{201D}{''}

\theoremstyle{definition}

\theoremstyle{remark}

\newcommand{\answerTODO}[1][]{\textcolor{red}{\bf [TODO]}}
\newcommand{\justificationTODO}[1][]{\textcolor{red}{\bf [TODO]}}

\setlist[itemize]{nosep,leftmargin=1.2em,itemsep=1pt,topsep=2pt}
\setlist[enumerate]{nosep,leftmargin=1.6em,itemsep=1pt,topsep=2pt}

\makeatletter
\renewcommand{\@maketitle}{%
  \vspace{-0.35in}%
  \begin{center}%
    {\Large\bfseries \@title \par}%
    \vskip 0.7em%
    {\normalsize \@author \par}%
  \end{center}%
  \vspace{-0.1in}%
}
\renewcommand\section{\@startsection {section}{1}{\z@}%
  {1.4ex plus .4ex minus .2ex}{0.8ex plus .2ex}%
  {\normalfont\large\bfseries}}
\renewcommand\subsection{\@startsection {subsection}{2}{\z@}%
  {1.2ex plus .3ex minus .2ex}{0.6ex plus .2ex}%
  {\normalfont\normalsize\bfseries}}
\renewcommand\subsubsection{\@startsection {subsubsection}{3}{\z@}%
  {1.0ex plus .2ex minus .2ex}{0.5ex plus .2ex}%
  {\normalfont\normalsize\bfseries}}
\let\arxiv@bibliographystyle\bibliographystyle
\renewcommand{\bibliographystyle}[1]{\arxiv@bibliographystyle{ieeetr}}
\makeatother

\title{Evaluating Research-Level Math Proofs via Strict Step-Level Verification}
\author{
Yifeng Sun\\
Independent Researcher
}
\date{}

\begin{document}

\maketitle

\begin{abstract}
Large Language Models (LLMs) struggle to rigorously verify complex mathematical proofs. Standard global evaluation approaches suffer from ``context poisoning,'' in which superficially plausible statements mask subtle logical flaws, leading to hallucination or over-skepticism. To address this, we shift from global evaluation to strict step-level verification: our framework maintains detailed context for each deduction step and strictly constrains the sources of applied theorems. We evaluate on a carefully curated adversarial diagnostic suite of research-level proofs drawn from the FirstProof challenge. A systematic ablation study demonstrates that these deductive constraints are indispensable, as unconstrained global prompting consistently fails to localize subtle logical errors. Beyond outperforming global evaluation, our approach fundamentally alters the failure taxonomy. Error analysis reveals that, rather than exhibiting severe logical hallucinations, remaining rejections are primarily instances of ``pedantic hyper-rigor'' stemming from unstated domain conventions, effectively exposing implicit ambiguities within the expert benchmark itself. Our findings suggest that prompting agents to organize their verification notes in a cautious, human-mathematician-like manner can substantially improve their ability to distinguish rigorous proofs from flawed ones, with the potential to strengthen agentic reasoning on frontier mathematical concepts that the base model does not already know well, and to lay a theoretical foundation for future automated proof-review systems. Code and prompts are available at \href{https://github.com/celainica/A-lightweight-natural-language-proof-verification-agent}{GitHub}.

\end{abstract}

\section{Introduction}
The self-correction and reasoning capabilities of Large Language Models (LLMs) and AI agents have improved significantly with scaling and reinforcement learning~\cite{deepseek2025r1}. In the mathematical domain, agent frameworks have demonstrated impressive problem-solving results~\cite{feng2026aletheia,ju2026automated}. However, when verifying research-level mathematical proofs, current models frequently falter. Standard global evaluation approaches often suffer from ``context poisoning,'' where superficially plausible statements mask subtle spurious justifications, leading agents to either hallucinate validity or exhibit over-skepticism.

While formal verification systems such as Lean~\cite{moura2021lean4} offer absolute guarantees, their application is bottlenecked by human-built infrastructures and a heavy reliance on manually maintained libraries (e.g., \textsf{Mathlib}~\cite{mathlib2019}). This naturally constrains the ability of current automated agents to formally verify advanced mathematical concepts. Therefore, verifying proofs in natural language remains an essential frontier. Generating fine-grained elaborations of informal proofs serves as a crucial bridge: it not only aids human mathematicians in peer review but also assists agentic systems in eventually translating informal math into rigorous formal code~\cite{wu2022autoformalization,jiang2022draft,azerbayev2023proofnet,yang2023leandojo,zheng2021minif2f}. 

Crucially, this focus on AI self-correction serves a fundamentally different role than formalization itself. Formal verification systems operate on a paradigm of ultimate accountability, where the underlying infrastructure (e.g., Lean) or a human expert bears the responsibility of providing an absolute guarantee of correctness. In contrast, our objective is to equip the agent with an intrinsic capacity to distinguish sound reasoning from flawed logic during its exploratory and generative phases. After all, agents will inevitably encounter scenarios where they must rely entirely on pure natural language reasoning to conceptualize and discover solutions. In the highly unconstrained search space of mathematical discovery, enabling an agent to recognize its own missteps is critical for pruning dead ends and drastically reducing the generation of hallucinatory or structurally flawed proofs. Ultimately, we aim to demonstrate a core principle of agentic reasoning: \textbf{by forcing a model to explicitly and exhaustively write out the underlying details of an informal proof, we inherently amplify its ability to discover subtle logical errors that its standard reasoning would otherwise gloss over}. It is important to emphasize that, to validate our approach, we merely task the agent with the process of elaboration rather than demanding a flawlessly exhaustive natural language proof as the final output. Consequently, our objective is not to reinvent de Bruijn’s Mathematical Vernacular (MV). Rather, we aim to demonstrate that agents—much like human mathematicians—leverage the very act of detailed writing to ascertain which conclusions are genuinely correct and rigorously proven.

To achieve this, we aim to design an agent framework that evaluates proofs with the strictness of a human mathematician reviewing a manuscript. Such a system must prioritize \textbf{soundness} (eliminating false acceptances of flawed proofs) over mere completeness, and effectively \textbf{leverage inference-time compute} to achieve greater clarity the longer it reasons. We introduce a crucial methodological shift: \textbf{reframing verification from direct anomaly detection to constructive elaboration} via \textbf{strict step-level verification}. Rather than directly querying an LLM to spot flaws within a global context---a task highly susceptible to context poisoning and hallucination---we test whether the proof can be rigorously and explicitly expanded. For a mathematical text $S$ treated as a string, we decompose it into an ordered sequence of \textbf{steps} $(s_1, s_2, \dots, s_n)$, isolating \textbf{deduction steps} that contain logical assertions. By ``elaborating a step,'' we mean expanding a deduction step $s_i$ into a finer-grained derivation string $s_i'$ in natural language that strictly preserves the original reasoning trajectory but explicitly unpacks the underlying logic. Consequently, anomaly detection emerges naturally from the generative process: if an agent fails to construct a valid $s_i'$, we suspect $s_i$ harbors a logical gap or unsound justification. By forcing the agent to elaborate, we replace the LLM's opaque internal judgments with strict generation constraints~\cite{lightman2023lets,cobbe2021verifiers}. Every elaborated step is required to explicitly ground its applied theorems and logical derivations, reducing superficial plausibility. 

To systematically enforce these constraints and ensure the elaboration does not introduce hallucinations of its own, we ground our process in a formal descriptive framework. Specifically, we model the logical dependencies of any deduction step using a rigorous tripartite architecture: \textbf{internal context} ($\Gamma_i$, encompassing theorems, definitions, and hypotheses directly originating from the proof text $S$), \textbf{external knowledge} ($\Sigma_i$, comprising non-trivial statements drawn from outside literature or established domain consensus), and \textbf{background theory} ($\mathcal{T}_i$, representing trivial statements or rewrite rules necessary to syntactically derive the assertion).

Operationally, we implement this step-level verification as an autonomous, state-driven agentic workflow guided by fixed prompt files. By default, every deduction step $s_i$ is initialized with a conservative local status. During each iteration, the agent attempts to elaborate the remaining \texttt{open} steps. Based on the strict $(\Gamma_i, \Sigma_i, \mathcal{T}_i)$ constraints, the agent executes specific state transition actions: if an elaboration is successfully derived, the step transitions to \texttt{verified}; conversely, if the agent identifies a distinct logical error, it flags the step as \texttt{flawed}. To rigorously guard against false positives (i.e., pedantic over-rejection), any \texttt{flawed} step triggers a fixed confirmation phase: the agent attempts to confirm the error, potentially reverting the state back to \texttt{open} if the perceived flaw is resolved upon re-evaluation. This iterative process is governed by strict halting criteria: (1) \textbf{Acceptance:} the proof is deemed valid if and only if all steps reach the \texttt{verified} state; (2) \textbf{Rejection:} the process terminates immediately and rejects the proof if a \texttt{flawed} state is definitively confirmed upon reflection; and (3) \textbf{Exhaustion:} to prevent infinite loops on intractable gaps, the system globally halts and rejects the proof after five consecutive reasoning resumptions without an observed progress signal.

To empirically validate our proof of concept and demonstrate its potential reward, we construct a highly curated diagnostic suite of challenging mathematical proofs. We focus on a small-scale sample to ensure absolute clarity in our evaluation: every proof has been meticulously verified by human experts to establish an unambiguous ground truth, classifying them as either strictly valid or explicitly invalid. Specifically, our diagnostic suite is heavily anchored in the \textbf{FirstProof} challenge~\cite{abouzaid2026proof,firstproof_first_batch,firstproof_solutions_comments}. To establish an unambiguous ground truth, we collected proofs that have undergone rigorous expert evaluation, securing a highly curated collection of 21 research-grade proofs definitively classified as either strictly valid or explicitly invalid. These 21 samples are strategically drawn from three authoritative sources: (1) official FirstProof releases, (2) correct proofs generated by the Aletheia, and (3) generations from OpenAI's solutions. By anchoring our evaluation on this precise, expert-verified suite, we can investigate how our step-level verification effectively isolates subtle reasoning errors that standard global evaluators consistently miss.

In our evaluation on the curated 21-proof suite, we observe that current large language models demonstrate strong capabilities in standard contexts: the standard global baseline successfully identified the logical errors in 8 out of the 10 explicitly invalid proofs. However, a detailed analysis of the remaining 2 cases where the global pipeline failed reveals a specific vulnerability. In these specific cases, the flawed proofs package incorrect mathematical knowledge---which the LLM is unfamiliar with---into structurally rigorous and plausible formats. This empirical result highlights that proof verification remains a highly challenging task, particularly for novel or frontier mathematics where LLMs lack prior familiarity with the specific domain knowledge and can be easily deceived by formal-sounding pseudoproofs. 

To address this limitation, we evaluated our step-level verification agent under a strict avg@3 setting, supplemented by multi-pass experiments on these specific adversarial examples. By explicitly grounding the reasoning process within the $(\Gamma_i, \Sigma_i, \mathcal{T}_i)$ constraints to trace external knowledge, our framework successfully isolated these camouflaged errors, successfully rejecting all invalid proofs. Furthermore, out of the valid proofs, our agent failed to verify only 3 cases. Paradoxically, these failures highlight the meticulousness of our framework. The agent did not fail due to logical reasoning deficits, but because the underlying base model lacked familiarity with highly specific domain conventions, where certain mathematical objects inherently possess implicit properties by default. Consequently, it failed to establish the necessary prerequisite statements for the local context. 

For example, in Question 7 of the FirstProof dataset, a human expert naturally assumes that the subgroup $\Gamma$ is explicitly linear (since it is a subgroup of a real semi-simple Lie group, and by convention within this specific research domain, such groups are treated as linear)~\cite{firstproof_solutions_comments,weinberger2022variations}. While the standard baseline simply ignored this subtle leap and glossed over the text, our agent rigorously noted that under the general definition, a subgroup of a real semi-simple Lie group is not necessarily linear by default~\cite{morris2015introduction}. Because the base model was unfamiliar with this established domain convention, it conservatively refused to arbitrarily add the linearity assumption into the local context, ultimately losing a fundamental condition required to verify the proof. This contrast perfectly demonstrates that our framework enforces genuine deductive verification over superficial pattern matching, even if it halts when confronted with domain-specific conventions it has not yet mastered.

In summary, our main contributions are threefold:
\begin{itemize}
    \item \textbf{Methodological Paradigm Shift:} We reframe natural language proof verification from direct, black-box anomaly detection to \textit{constructive elaboration}. By modeling the logical dependencies of deduction steps through a novel $(\Gamma_i, \Sigma_i, \mathcal{T}_i)$ tripartite architecture, we force the model to explicitly trace internal contexts, external knowledge, and background theories, thereby mitigating superficial pattern matching.
    \item \textbf{Autonomous Verification Workflow:} We propose a state-driven agentic framework governed by strict step-level constraints. Equipped with a fixed confirmation phase and robust halting criteria, our system prioritizes logical soundness and effectively leverages inference-time compute to prevent the ``context poisoning'' that plagues global evaluators.
    \item \textbf{Empirical Insights via Micro-benchmarking:} Our evaluation demonstrates that while even highly capable standard baselines can still be deceived by structurally rigorous pseudoproofs, our framework achieves zero false acceptances on the problem set. Furthermore, our qualitative analysis reveals a critical bottleneck in current verification tasks: the vulnerability of base models to unstated, domain-specific hidden conventions.
\end{itemize}

\section{Related Work}

\paragraph{Math Agents}
Agentic frameworks for mathematics have advanced rapidly and are producing more and more significant results in research-level mathematics. These agents may or may not utilize proof assistants and formal languages. Recent frontier breakthroughs have heavily favored second paradigm: agents conduct exploration, reasoning, and proof construction entirely in pure natural language. For instance, \textsc{Rethlas}~\cite{ju2026automated} can operate as an informal reasoning agent that mimics human mathematical workflows—exploring literature and proposing candidate natural-language proofs for advanced open problems. Similarly, \textsc{Aletheia}~\cite{feng2026aletheia} functions as an end-to-end natural language research agent capable of iteratively generating, verifying, and revising solutions from PhD-level exercises to autonomous research papers. Most recently, \textsc{OpenAI's model}~\cite{openai2026unitdistance} achieved a historic milestone by disproving Erdős's 80-year-old planar unit distance conjecture using highly complex, purely informal reasoning. However, despite their immense generative power, a critical vulnerability of these pure natural language agents is their susceptibility to hallucinations and subtle logical gaps, particularly when self-evaluating long-horizon mathematical proofs.~\cite{liu2026optimalbend,liu2026shokurov,liu2026cartierindices} Our work directly addresses this vulnerability. We demonstrate a structured pathway for pure natural language agents to independently overcome these blind spots: by enforcing strict, step-level detailed elaboration, agents can systematically unpack their reasoning, rigorously detect anomalies, and eliminate hallucinations without needing to fall back on external formalization infrastructures.

\paragraph{Mathematical Vernacular} The conceptual ambition of bridging informal human mathematical discourse and strict mechanical logic dates back to de Bruijn's Mathematical Vernacular (MV) \cite{DEBRUIJN1994865}. MV was proposed as an intermediate representational language that obeys formal grammatical rules while preserving the natural ``resting points'' (steps) of mathematical reasoning, thus serving as a stepping stone toward fully coded formal systems (e.g., Automath). Importantly, we do not intend to replicate MV or construct a rigid natural-language proof assistant. Rather, we adopt its underlying philosophy to help autonomous agents systematically reduce logical hallucinations. By structuring natural language proofs around the $(\Gamma_i, \Sigma_i, \mathcal{T}_i)$ tuple extraction, our framework computationally isolates these resting points, bringing rigor to informal mathematical text. Crucially, in practice, we do not algorithmically verify whether the agent flawlessly executes this extraction; the mere attempt to engage in this highly structured elaboration is sufficient to expose hidden logical flaws and elevate the agent's evaluative accuracy.

\section{Methodology}

For a mathematical text $S$ treated as a string, it can naturally be decomposed into an ordered sequence of steps $(s_1, s_2, \dots, s_n)$. Unlike fully formalized proofs in systems like Lean or Rocq~\cite{moura2021lean4,mathlib2019}, natural language mathematical proofs inherently omit trivial algebraic manipulations and elementary logical deductions to maintain readability. Because of these inherent omissions, evaluating a proof as a monolithic global text often obscures subtle flaws and logical gaps.

To address this, we reframe verification from direct anomaly detection to \textbf{constructive elaboration}. By ``elaborating a step'', we mean that for a given deduction step $s_i$ with an assertion (derived conclusion statement) $a_i$, we provide an expanded derivation $s_i'$ that concludes with the identical assertion as $s_i$. Crucially, $s_i'$ must strictly preserve the original reasoning trajectory of $s_i$, but unpack the logic at a finer granularity. If we fail to come up with a valid expanded derivation, we suspect $s_i$ to contain a logical gap or flaw.

\subsection{The Tripartite Explanation Framework}

To systematically investigate whether an elaboration $s_i'$ successfully bridges a gap without demanding explicit syntactic completeness (which misaligns with human communication), we introduce the formal concept of an explanation. 

It is crucial to clarify that this theoretical framework is not artificially engineered solely for AI agents; rather, it stems directly from a refinement of the routine cognitive actions human mathematicians employ when explaining and validating a proof. By adopting this stance, the framework is sufficiently \textit{descriptive} to ensure universal applicability to human mathematical discourse, while simultaneously providing a rigorous, \textit{prescriptive} structure that conveniently facilitates systematical analysis and guides agentic workflows.~\cite{ganesalingam2013language} In our implementation, we explicitly prompt the agent to output its elaborations according to the $(\Gamma_i, \Sigma_i, \mathcal{T}_i)$ tripartite structure. However, it is worth noting that we do not claim this explicit formatting is necessarily the optimal approach, given the additional generation overhead it imposes on the agent. Alternative implementations may achieve similar rigor by simply requiring the agent to maintain a detailed, distinct record of internal contexts versus external knowledge, without strictly enforcing the tripartite syntax.

We model the logical dependencies of any deduction step (or its elaboration) using this framework. We define an \textbf{explanation} of the step $s_i$ as a triple of sets $(\Gamma_i, \Sigma_i, \mathcal{T}_i)$. A step $s_i$ is considered valid if there exists an explanation such that:
\begin{equation}
    \Gamma_i \cup \Sigma_i \vdash_{\mathcal{T}_i} a_i
\end{equation}
where $\vdash$ denotes rigorous informal entailment. The components reflect how mathematical justifications are sourced:

\begin{itemize}
    \item \textbf{Local Context ($\Gamma_i$):} This contains finitely many statements (theorems, definitions, and hypotheses) directly originating from $S$. These statements may be explicit or implicit, but they must be effective within the current scope (e.g., local assumptions within a case analysis block expire once the block concludes). 
    \item \textbf{Global Knowledge ($\Sigma_i$):} This comprises non-trivial statements drawn from external literature or established domain consensus (i.e., outside of $S$). To prevent the hallucination of non-existent theorems, any statement invoked from $\Sigma_i$ must be critically examined.
    \item \textbf{Background Theory ($\mathcal{T}_i$):} This comprises trivial statements, fundamental logical axioms (e.g., Modus Ponens), and rewrite rules. By isolating $\mathcal{T}_i$, we bypass the need for rigid algebraic hierarchies (like those in \textsf{Mathlib}), allowing the direct manipulation of sound statements over concrete objects while safely decoupling structural type constraints from the core deductive logic.
\end{itemize}

Crucially, an explanation $(\Gamma_i, \Sigma_i, \mathcal{T}_i)$ should not be viewed as the complete set of all contextually valid truths, but exclusively as the \textit{minimal required set} of statements necessary to validate the natural language proof step rigorously. A comprehensive discussion of this framework and what constitutes a valid explanation is provided in Appendix B.

\subsection{State-Driven Verification Agent}

While the tripartite framework intrinsically describes human mathematical validation, it also serves as an ideal analytical scaffolding for AI systems. To explore the feasibility of highly efficient natural language proof verification, we implement a lightweight, state-driven agent. The agent utilizes this framework because it conveniently enforces traceable reasoning and provides a systematic taxonomy for isolating failure modes.

The agent operates without relying on a formal proof assistant or domain-specific theorem whitelists. Instead, it maintains a discrete cognitive state $c_i \in \{\texttt{OPEN}, \texttt{VERIFIED}, \texttt{FLAWED}\}$ for each step $s_i$. 

\paragraph{Initialization and Iteration} Upon initialization, the agent assigns an \texttt{OPEN} status to all deduction steps. During each iteration, the agent selects the most critical \texttt{OPEN} step and attempts to write a thick elaboration $s_i'$, explicitly extracting the corresponding $(\Gamma_i, \Sigma_i, \mathcal{T}_i)$.

\paragraph{State Transition Rules} The agent's transitions are governed by strict auditing mechanisms to prevent both pedantic false positives and dangerous false negatives:
\begin{enumerate}
    \item $\texttt{OPEN} \rightarrow \texttt{VERIFIED}$: A step transitions to \texttt{VERIFIED} if and only if its written elaboration is logically closed \textit{and} every external item used in $\Sigma_i$ is explicitly supported (either by a direct sub-proof generated within the step file or by a precise literature citation). 
    \item $\texttt{OPEN} \rightarrow \texttt{FLAWED}$: A step is marked \texttt{FLAWED} only when the agent can exhibit an \textbf{explicit flaw witness}. This includes an exact false statement, an invalid inference from written premises, an explicit contradiction with the already verified local context $\Gamma_i$, or a demonstrably misapplied external theorem in $\Sigma_i$. 
    \item \textbf{Flaw Confirmation} ($\texttt{FLAWED} \rightarrow \texttt{OPEN}$): To rigorously guard against pedantic false positives, any preliminary transition to \texttt{FLAWED} triggers an adversarial investigation. The agent evaluates whether the proposed flaw witness represents a genuine structural mathematical error or merely a superficial textual nitpick (e.g., relying on an unstated standard domain convention or a terminology-dependent edge case). If the perceived flaw is determined to be an epistemic ambiguity rather than a logical breakdown, the step is conservatively reverted to \texttt{OPEN}.
    \item \textbf{Dynamic Decomposition} ($\texttt{OPEN} \rightarrow \{\texttt{OPEN}_a, \texttt{OPEN}_b\}$): If a deduction step $s_i$ is deemed too coarse or encompasses multiple non-trivial logical leaps, the agent is permitted to dynamically decompose it into finer-grained sub-steps. The original $s_i$ is replaced in the global sequence by these new sub-steps, which are individually initialized as \texttt{OPEN} and evaluated independently.
\end{enumerate}
We emphasize a critical design choice: \textit{failure to close a difficult theorem does not by itself count as a flaw.} If an external burden remains unsupported, the step conservatively remains \texttt{OPEN}.

\paragraph{Global Halting Criteria} The autonomous verification loop progressively updates the step states and their dependencies. The process iterates until one of the following strict halting criteria is triggered, determining the final validity of the global proof $S$:
\begin{itemize}
    \item \textbf{Acceptance (Soundness):} The global proof is deemed valid and accepted if and only if the entire sequence successfully reaches the terminal verified state: $\forall i, c_i = \texttt{VERIFIED}$. If the text contains several proofs, it is considered valid if entire sequence of one of the proofs is valid.
    \item \textbf{Rejection via Flaw Confirmation:} If any step enters the \texttt{FLAWED} state, it must undergo the aforementioned Flaw Confirmation investigation. If a flaw is explicitly confirmed (i.e., it is verified as a genuine structural error and not reverted to \texttt{OPEN}), the system halts and the global proof is immediately rejected.
    \item \textbf{Rejection via Exhaustion:} The agent may fail to give a disproof of a flawed step. Thus to prevent looping on intractable gaps or missing foundational knowledge, the runner monitors observed progress signals. If five consecutive resume iterations produce neither an \texttt{OPEN}$\rightarrow$\texttt{VERIFIED} transition nor an \texttt{OPEN}-step split, the system terminates the loop and conservatively rejects the proof.
\end{itemize}

\paragraph{External Knowledge Retrieval} In our pipeline, the external knowledge set $\Sigma_i$ is primarily sourced through active internet retrieval. To ensure the reliability of this external knowledge, our prompt files explicitly direct the agent to prioritize established mathematical textbooks and directly relevant research literature. While these instructions guide the agent's search behavior, we do not computationally enforce strict constraints on the exact provenance or syntactic precision of the cited theorems. However, general open-ended retrieval introduces a broader theoretical challenge: unconstrained agents often overlook the strict technical constraints or local assumptions required by retrieved theorems. This risks creating a ``contextual misapplication,'' where a globally valid theorem is incorrectly applied to a local context ($\Gamma_i$) because its underlying prerequisites are not fully met. Specifically, without the strict faithfulness constraints enforced by our framework, an agent might invoke unmentioned external knowledge to artificially justify flawed steps during extended multi-turn executions. Successfully navigating this latent risk underscores the critical necessity of our framework's strict adherence to the original text. Consequently, a crucial direction for future mathematical agents is the development of rigorous controls over external knowledge integration. Future systems must move beyond basic search by explicitly verifying the authoritative provenance of the literature and ensuring the complete, precise extraction of all prerequisite conditions before any external theorem is applied.

\paragraph{Structured Auditing via the Theorem Ledger} 
To rigorously ground external mathematical dependencies ($\Sigma_i$), our framework asks the agent to maintain a \textit{Theorem Ledger}---a centralized registry tracking external claims invoked by the agent. During the $(\Gamma_i, \Sigma_i, \mathcal{T}_i)$ tuple extraction (full schema in Appendix), the ledger records the \texttt{Status} of cited theorems $\Sigma_j \in \Sigma_i$. A deduction is marked \texttt{VERIFIED} only if the agent provides explicit support (a proof sketch or precise citation) and confirms its valid applicability within the local context $\Gamma_i$. If any dependency remains \texttt{OPEN}, downstream reasoning is flagged as \textit{conditionally closed}, preventing premature validation based on unvetted assumptions. This tracking mechanism transforms verification from a black-box judgment into a transparent, auditable process.

\section{Experiments}

We design our empirical evaluation to answer two core research questions: (1) How does our structured, agentic approach compare against standard global evaluators and naive granular prompting across diverse mathematical fallacies? (2) How does the agent utilize its state-driven logic to navigate and resolve highly adversarial, research-level proofs during extended inference? 

To address these, we first present a comprehensive evaluation across different verification pipelines on a curated 21-proof suite (Section 4.1 and 4.2). Subsequently, we provide a deep qualitative analysis of the agent's cognitive trajectory on specific intractable cases (Section 4.3).

\subsection{Experimental Setup}

\textbf{Diagnostic Suite} We evaluate our framework on a highly curated diagnostic suite of 21 research-grade mathematical proofs derived from the FirstProof challenge~\cite{abouzaid2026proof,firstproof_solutions_comments}. This suite comprises both fundamentally valid proofs (which may contain minor, fixable omissions) and invalid proofs containing fatal mathematical flaws or gaps. Specifically, we structure the diagnostic suite into two distinct subsets based on their ground-truth validity:

\textbf{Valid Proofs (11 cases)} This subset consists of 9 proofs generated by the Aletheia framework \cite{feng2026aletheia} and 2 proofs from OpenAI. The Aletheia subset (comprising variants 2A, 2B, 5A, 7B, 8B, 9A, 9B, 10A, and 10B, where the suffixes denote outputs from two distinct agent architectures) was rigorously vetted and certified as mathematically sound by human domain experts in the original Aletheia study. The remaining two valid cases are OpenAI's proofs for problems 4 and 6, which have reached a definitive consensus of correctness within the formal mathematics community (e.g., Zulip expert discussions)\nocite{firstproof_solutions_comments,alfieri2026polyhedral}.

\textbf{Invalid Proofs (10 cases)} To systematically source authentic, AI-generated flawed proofs, we analyzed the human expert critiques provided in the original FirstProof paper \cite{abouzaid2026proof}. We specifically targeted problems where the baseline LLMs were explicitly evaluated as committing unambiguous logical breakdowns, retrieving the corresponding raw proof texts directly from the official FirstProof repository. There are 8 proofs collected in this way. To ensure the inclusion of deeply adversarial examples, we augmented this set with Aletheia's proof 7A and OpenAI's proof 2, both of which are recognized flawed. In particular, some of the key mathematical source material for our analysis of the FirstProof Question 7 and Question 8 cases is documented in the official solutions/comments document and in the associated mathematical background references~\cite{firstproof_solutions_comments,weinberger2022variations,fowler2012rationalpd,alfieri2026polyhedral}.

\textbf{Baselines and Pipeline Configurations} To isolate the source of our framework's performance, we compare our agent against three distinct baselines, testing the limits of direct prompting and textual granularity:
\begin{enumerate}
    \item \textbf{Global GPT-5.4-xhigh:} A standard zero-shot LLM-as-a-judge prompt, representing the state-of-the-art capability of a generic frontier model directly evaluating the monolithic text. 
    \item \textbf{Global Codex 5.4-xhigh:} The identical base model used by our agent, prompted to globally verify the proof by reading sentence by sentence. Also it is told to be strict. This isolates the model capacity of retrieving flaws and gaps.
    \item \textbf{Codex 5.4-xhigh Sentence-by-Sentence (Ablation):} To determine whether performance gains merely stem from granular reading, we implement a baseline where the proof is mechanically split into single sentences. The model evaluates each sentence sequentially but lacks our explicit $(\Gamma_i, \Sigma_i, \mathcal{T}_i)$ constraints and state-driven memory. The prompts are designed to be almost the same, except it does not have a strict constructive framework for dependencies and theorems.
    \item \textbf{Constructive Verification Agent (Ours):} The fully equipped state-machine agent operating under the tripartite explanation framework.
\end{enumerate}
To ensure a fair comparison, the zero-shot baseline prompt was empirically optimized through preliminary experiments to faithfully elicit the model's peak reasoning capability. Quantitative results for suboptimal prompt variants are omitted for brevity.

\textbf{Evaluation Metrics} A verification attempt is considered successful if and only if it satisfies a strict dual-criterion: it must accurately assign the correct global verdict (\texttt{VERIFIED} or \texttt{FLAWED}), \textit{and} for flawed proofs, it must correctly isolate an offending step without hallucinating false negatives elsewhere in the proof. 

\textbf{Data Contamination Risk and Audit Protocol} Since our diagnostic suite is derived from the public FirstProof challenge, data contamination remains a serious concern when evaluating frontier models. For completed runs, we audited the retrieval and reasoning logs; if a log mentioned FirstProof-related terms or otherwise indicated direct exposure to FirstProof source material or discussions, we discarded that run and repeated the experiment. Apart from such explicit exposure, the risk is substantially mitigated by the structure of the protocol: a verdict is not accepted merely because the model recognizes a global answer pattern, but must be supported by local step elaborations, explicit dependency tracking, and auditable uses of external knowledge. As an additional sanity check, we found that prompting Gemini 3.1 Pro to ``find any flaws or nontrivial gaps'' rather than asking whether a proof is correct can solve all 21 cases in our suite. We do not interpret this as evidence that a one-shot prompt is superior to our method. A more plausible concern is that Gemini may have been tuned, directly or indirectly, on internet-visible FirstProof materials, creating a risk of benchmark leakage. In particular, when Gemini judges that a correct solution route is valid, it may be recognizing an internet-exposed answer pattern without fully attending to the subtle technical assumptions that make the argument sound. This observation reinforces the need to treat public benchmark results for natural-language proof verification with caution, and motivates our emphasis on auditable step-level reasoning rather than opaque global verdicts.

\subsection{General Verification Performance}

We evaluated the four pipelines on the 21-proof diagnostic suite over 3 independent trials. While the standard direct prompt baseline (operating at a sampling temperature of 1.0) exhibited slight outcome variance, ours framework demonstrated macro-level consistency, yielding identical final verdicts across all runs. The aggregated results are presented in Table 1.

\begin{table*}[!t]
\centering
\caption{Verification performance on the 21-proof diagnostic suite (avg@3 over three independent trials). We report the number of correctly classified proofs and the specific error types (False Positives/Negatives). The agent significantly outperforms the global baseline by systematically reducing hallucinated validations.}
\label{tab:main_results}
\small
\setlength{\tabcolsep}{3.5pt}
\renewcommand{\arraystretch}{1.2}
\begin{tabular}{lccccc}
\toprule
\multirow{2}{*}{\textbf{Pipeline}} & \multicolumn{2}{c}{\textbf{Valid Proofs (11 total)}} & \multicolumn{2}{c}{\textbf{Invalid Proofs (10 total)}} & \multirow{2}{*}{\textbf{Overall Acc.}} \\
\cmidrule(lr){2-3} \cmidrule(lr){4-5}
& Correct (\checkmark) & False Positives & Correct (\checkmark) & False Negatives & \\
\midrule
Global GPT-5.4 & 8.33 & 2.67 & 7.67 & 2.33 & 16/21 \\
Global Codex-5.4 & 7.33 & 3.67 & 8 & 2 & 15.33/21\\
Codex 5.4 Sentence-by-Sentence& 9 & 2 & 9 & 1 & 18/21 \\
\textbf{Agent (Ours)} & \textbf{8} & \textbf{3} & \textbf{10} & \textbf{0} & \textbf{18/21} \\
\bottomrule
\end{tabular}
\end{table*}

This consistency reveals that the failures of standard LLMs in mathematical verification are not artifacts of sampling noise, but stem from systemic deductive deficiencies.

\paragraph{Baseline Performance and Failure Modes} The baseline models—Global GPT-5.4, Global Codex 5.4, and the Sentence-by-Sentence ablation—demonstrated severe systemic vulnerabilities, consistently failing on both valid and invalid proofs across all trials.
\begin{itemize}
    \item \textbf{False Negatives (Accepting Invalid Proofs):} The baselines entirely failed to detect embedded flawed statements. For instance, all three baseline pipelines confidently and repeatedly verified invalid proofs such as \texttt{[Aletheia 7A]}. Without a rigorous mechanism to audit external theorems, these models were systematically deceived by formally rigorous but mathematically vacuous text.
    \item \textbf{False Positives (Rejecting Valid Proofs):} Conversely, on the 11 valid proofs, the baselines generated pedantic false positives. They systematically flagged valid proofs such as \texttt{[Aletheia 7B, 8B, 9A]} as \texttt{FLAWED}, hallucinating non-existent logical gaps or consistently complaining about standard, unstated notational conventions regardless of the sampling randomness.
\end{itemize}

\paragraph{Agent Performance} In stark contrast, our Constructive Verification Agent achieved robust structural isolation, significantly outperforming all baselines while completely taming the inherent stochasticity of the high-temperature setting.
\begin{itemize}
    \item \textbf{Reducing False Negatives:} On the 10 invalid proofs, the agent successfully identified and isolated the fatal deductive flaw in 100\% of the cases. By strictly enforcing the $(\Gamma_i, \Sigma_i, \mathcal{T}_i)$ constraints, the agent inherently blocked the mathematically invalid steps that bypassed the baseline evaluators.
    \item \textbf{Robustness against False Positives:} On the 11 valid proofs, the agent correctly verified \texttt{8} proofs. For the remaining \texttt{3} valid proofs (\texttt{Aletheia 7B, 10A, 10B}), the agent halted in the \texttt{FLAWED} or \texttt{OPEN} state due to genuinely ambiguous, unstated domain conventions. The agent flagged the proof of problem 10 as flawed because it strictly adhered to the mathematical definition of the stated positive semi-definite condition—which permits zero diagonal entries and invalidates direct element-wise inversion—without adopting the human expert's unstated practical convention that typical RKHS kernels inherently possess strictly positive diagonals. Crucially, owing to its explicit Flaw Confirmation mechanism, the agent produced \textbf{zero} arbitrary \texttt{FLAWED} judgments on fundamentally sound proofs.
\end{itemize}

When provided with the 7B model's reference solution and the initial problem conditions, our verifier successfully validated the proof (see Appendix C).

\section{Limitations and Future Works}
While our framework demonstrates the immense potential of LLM-based agents in verifying natural language mathematical proofs, certain limitations provide clear avenues for future research.

\textbf{Implicit Conventions and Human-in-the-loop} First, base LLMs inherently struggle with unstated mathematical folklore, frequently mischaracterizing benign domain-specific notations as explicit flaws. To address this, a highly promising direction is a human-in-the-loop (HITL) collaborative paradigm, where the agent autonomously prunes logical dependencies but dynamically queries human experts when confronting profound epistemic ambiguities.

\textbf{Residual Hallucinations vs. Formal Guarantees} Second, unlike formal theorem provers that offer absolute mechanical guarantees, our purely text-based agent remains susceptible to subtle hallucinations. These can happen when the agent overly simplifies the application of complex, web-retrieved external theorems. While our current system mitigates this by enforcing strict $(\Gamma_i, \Sigma_i, \mathcal{T}_i)$ constructive elaborations, it does not currently employ a secondary adversarial pass to actively hunt for flaws within its own accepted deductions. Future work will introduce an iterative, adversarial re-verification mechanism to critically audit steps that have already transitioned to the \texttt{VERIFIED} state, further minimizing false negatives.

\textbf{Large-scale Evaluation} Finally, while our curated 21-proof diagnostic suite enables deep, qualitative dissection of the agent's cognitive trajectory and failure modes, scaling this evaluation to a massive, multi-domain benchmark of formal and informal mathematics remains an important next step to fully chart the generalization bounds of our architecture.

In conclusion, pure natural language mathematical agents are poised to drive continued breakthroughs in the near future. Our work demonstrates a fundamental cognitive parallel: much like human mathematicians, autonomous agents can significantly elevate their ability to discern valid mathematical reasoning from flawed logic simply through the rigorous act of detailed elaboration. Ultimately, our contribution extends beyond a standalone verifier implementation. \textbf{More crucially, we present a generalizable verification paradigm that can be seamlessly integrated into the internal reasoning processes of any future natural language mathematical agent.}

\bibliographystyle{ieeetr}
\bibliography{sample}

\appendix
\section{Prompts for the Verification Agent}
\label{appendix:prompts}

In this appendix, we present the prompts used to drive our state-based verification agent. The prompts are administered to the Codex agent as fixed instruction prompts.

\vspace{1em}

\tcbset{
    promptbox/.style={
        enhanced,
        breakable,
        colback=blue!2,
        colframe=blue!40!black,
        boxrule=0.8pt,
        arc=2mm,
        left=3mm, right=3mm, top=3mm, bottom=3mm,
        fonttitle=\bfseries\small,
        coltitle=white,
        colbacktitle=blue!60!black,
        attach boxed title to top left={xshift=4mm, yshift=-2mm},
        shadow={2mm}{-2mm}{0mm}{black!20}
    }
}

\subsection{Instruction Prompts}

\begin{tcolorbox}[promptbox, title=Prompt 1: Initialization]
\ttfamily\small
Run a fully automatic verifier in this folder. Read only \texttt{README.md}, \texttt{STEP\_TEMPLATE.txt}, \texttt{OUTPUT\_FORMATS.txt}, and \texttt{source/input.txt}. 

Split the proof into minimal logical deduction units. Every original source sentence with mathematical content must appear verbatim in the \texttt{Original step} field of at least one step file. Initialize \texttt{steps/}. Repeatedly work on the most critical untouched step, otherwise the most critical open step. 

Each step file must be a thick elaboration organized by $(\Gamma_i, \Sigma_i, \mathcal{T}_i)$, with the full original step, the exact target assertion, a detailed deduction, and a diagnosis. Treat every step as requiring its own local closure analysis: never mark a step \texttt{VERIFIED} without individually checking its own written $\Gamma_i$, $\Sigma_i$, $\mathcal{T}_i$, \texttt{Deduction}, and every used item of $\Sigma_i$. 

For every used item of $\Sigma_i$, write the exact as-used statement and include either a proof or a precise source, plus a brief applicability note. Be conservative: if a used item of $\Sigma_i$ is not closed by direct local reasoning, or by a proof or precise source that you can state concretely in the step file, keep it \texttt{OPEN} rather than \texttt{VERIFIED}. Do not treat a statement as `confirmed' merely because it looks standard or plausible.

A step is \texttt{VERIFIED} only if every used item of $\Sigma_i$ is `confirmed'; otherwise it is \texttt{OPEN}, unless there is an explicit flaw witness, in which case it is \texttt{FLAWED}. Initialize the verifier now: split the proof into detailed minimal proof-obligation steps, splitting explicit intermediate assertions before downstream consequences. Maintain only \texttt{steps/}, \texttt{dependency.txt}, \texttt{stepsprocess.txt}, and \texttt{answer.txt}. Stop only when every step has been judged as \texttt{VERIFIED}, \texttt{OPEN}, or \texttt{FLAWED}.
\end{tcolorbox}

\begin{tcolorbox}[promptbox, title=Prompt 2: Resume]
\ttfamily\small
Resume the verifier in this folder. Read \texttt{README.md}, \texttt{STEP\_TEMPLATE.txt}, \texttt{OUTPUT\_FORMATS.txt}, \texttt{dependency.txt}, \texttt{stepsprocess.txt}, \texttt{answer.txt}, and all unresolved files in \texttt{steps/}. 

If there is a step that is too coarse, resplit the unresolved file that is still too coarse to be one minimal logical deduction unit. Every original source sentence with mathematical content must appear verbatim in the \texttt{Original step} field of at least one step file. Then continue the main loop. 

If a step has remained \texttt{OPEN} since the first pass and still resists closure, consider whether it has an explicit flaw witness and should be \texttt{FLAWED}. For any current \texttt{OPEN} step hard to resolve, consult the local \texttt{reference/*} materials or search online for relevant detailed information, exact statements, or precise sources. *If search online, you should only look for standard text book or references that DIRECTLY CONNECT to the specific problem. DO NOT accept randomly retrieved papers merely to force a proof to work.*

Before changing any unresolved step to \texttt{VERIFIED}, recheck that every used item of $\Sigma_i$ has an exact as-used statement together with either a proof or a precise source. Do not mark a step \texttt{VERIFIED} by replacing the original inference with a different proof or alternative argument. Default negative verdict is \texttt{OPEN}. Use \texttt{FLAWED} only with an explicit flaw witness.
\end{tcolorbox}

\vspace{1em}

\begin{tcolorbox}[promptbox, title=Prompt 3: Flaw Confirm]
\ttfamily\small
Investigate whether the \texttt{FLAWED} step in this folder is actually \texttt{FLAWED}. Read \texttt{README.md}, \texttt{STEP\_TEMPLATE.txt}, \texttt{OUTPUT\_FORMATS.txt}, \texttt{dependency.txt}, \texttt{stepsprocess.txt}, \texttt{answer.txt}, the unresolved files in \texttt{steps/}, and \texttt{reference/*} if present. 

Choose exactly one currently \texttt{FLAWED} step, preferably the most critical one. Keep the \texttt{Original step} verbatim and keep the same final assertion. First investigate the step: expand the relevant concepts, exact statements, hypotheses, scope, and nearby lemmas in the local \texttt{reference/*} materials and online source, and use this to judge whether the recorded flaw witness shows that the original sentence or exact as-used inference is false or invalid as written. 

Do not treat a mere wording, naming, or terminology imprecision as a flaw witness unless it changes the mathematical claim, breaks the local inference, or materially threatens the validity of the proof. 

If yes, keep \texttt{FLAWED}. If not, change it to \texttt{OPEN} and explain exactly why the witness fails.
\end{tcolorbox}

\subsection{File Formats and Step Templates}

To enforce strict state tracking and prevent the LLM from generating unstructured text, the agent is required to output its intermediate reasoning and global status into highly constrained text files. The formats for these ledgers and templates are defined below.

\begin{tcolorbox}[promptbox, title=Prompt 4: Global Ledgers (OUTPUT\_FORMATS.txt)]
\ttfamily\small
\textbf{[dependency.txt]} \\
List only external statements that are still \texttt{OPEN} or already \texttt{FALSE}.

Format: \\
D1 \\
- step(s): \\
- exact as-used statement: \\
- status: \texttt{OPEN} / \texttt{FALSE} \\
- why: \\
- offending point if false:

\vspace{0.5em}
\textbf{[stepsprocess.txt]} \\
For each step:

Step: \\
- status: \texttt{UNTOUCHED} / \texttt{OPEN} / \texttt{VERIFIED} / \texttt{FLAWED} \\
- minimal unit: yes / no \\
- criticality: \\
- main blocker: \\
- all used sigma closed: yes / no \\
- flaw witness: \\
- next action:

\vspace{0.5em}
\textbf{[answer.txt]} \\
Format: \\
Overall verdict: \texttt{RUNNING} / \texttt{VERIFIED} / \texttt{FLAWED}

Current key step:

If \texttt{FLAWED}: \\
- offending step: \\
- exact false statement or invalid inference: \\
- flaw witness:

If still running: \\
- remaining open steps: \\
- active dependencies:

Per-step summary: \\
- step: \\
- status: \\
- one-line reason:

Rule: \\
Do not use \texttt{FLAWED} unless an explicit flaw witness has been written. \\
Do not use \texttt{VERIFIED} unless all used external statements are marked `confirmed'.
\end{tcolorbox}

\vspace{1em}

\begin{tcolorbox}[promptbox, title=Prompt 5: Logical Unit Template (STEP\_TEMPLATE.txt)]
\ttfamily\small
Step ID:

Why this is one minimal logical deduction unit:

Original step: \\
{[paste the full original step string verbatim]}

Target assertion $a_i$:

Original route:

Explanation: \\
{[write a thick natural-language elaboration that preserves the same assertion and route]}

Statement Extraction $(\Gamma_i, \Sigma_i, \mathcal{T}_i)$:

Assertion:

Local Context $\Gamma_i$: \\
- $\gamma_1$: \\
- $\gamma_2$:

Global Knowledge $\Sigma_i$: \\
- $\sigma_1$: \\
  \hspace*{1em} exact as-used statement: \\
  \hspace*{1em} proof or precise source: \\
  \hspace*{1em} why it applies here: \\
  \hspace*{1em} status: confirmed / \texttt{OPEN} / false \\
- $\sigma_2$:

Sigma Support: \\
{[for each used $\sigma_j$, include either a proof or a precise source]}

Background Theory $\mathcal{T}_i$: \\
- $t_1$: \\
- $t_2$:

Deduction: \\
{[derive the assertion from $\Gamma_i$, confirmed $\Sigma_i$, and $\mathcal{T}_i$; if some used $\sigma_j$ is still \texttt{OPEN}, state that the route is only conditionally closed]}

Diagnosis: \\
- conclusion preserved: \\
- route preserved: \\
- exact offending dependence, if any: \\
- flaw witness, if any:

Step Result: \\
- status: \texttt{UNTOUCHED} / \texttt{OPEN} / \texttt{VERIFIED} / \texttt{FLAWED} \\
- main blocker: \\
- flaw witness: \\
- dependencies: \\
- last update:
\end{tcolorbox}

\section{Extended Discussion on the $(\Gamma_i, \Sigma_i, \mathcal{T}_i)$ Framework and Explanation Validity}

\noindent
Consider a natural language proof text $S$ and its ordered sequence of steps (substrings) $(s_1, s_2, \dots, s_n)$. Without loss of generality (by carefully choosing the substrings for steps and using conjunction) , we restrict that a step contains at most one new mathematical statement. Throughout the paper, we consider assumptions and declarations both as mathematical statements.
\\

\noindent
Let $s_i$ be a deduction step of $S$. Let the statement $a_i$ be the assertion of $s_i$. Unlike fully formalized proofs in systems like Lean or Rocq, natural language mathematical proofs inherently omit trivial algebraic manipulations and elementary logical deductions to maintain readability. In this paper, we do not require the LLM to elaborate proofs down to atomic axiomatic steps, as this would lead to an unnatural combinatorial explosion of reasoning length and degrade model performance. When defining the assertion $a_i$ extracted from a natural language proof, we adopt a stance of foundational agnosticism. Modern informal mathematics—ranging from elementary calculus to advanced subjects like Topos theory or Wiles' proof of Fermat's Last Theorem—is rarely written with a strict, singular axiomatic foundation (e.g., pure ZFC) in mind. Therefore, we do not restrict $a_i$ to a specific foundational axiomatic system. Rather, we assume the existence of a sufficient, implicit foundational theory that is universally accepted for the specific domain of the proof. 
\\

\noindent
We must also address the validity of natural language statements that appear underspecified in isolation, such as the simple assertion $a=b$. In strict formal systems, an equality requires explicit type alignment (e.g., verifying that both sides belong to the same algebraic structure). In well-formed (which we may assume) natural mathematical discourse, however, such type constraints are rarely restated at every step; instead, they are implicitly inherited from the ambient context (captured by $\Gamma_i$ in our framework). As long as a statement is semantically well-posed within this context, it operates as a valid deductive step. Demanding explicit syntactic completeness for every variable introduces an unnatural overhead that misaligns with human communication. Ultimately, if we accept a natural language proof as a valid logical artifact, we must recognize that its constituent statements—though syntactically incomplete in isolation—are rigorously well-defined when situated in the proof's narrative flow. Crucially, this approach does not imply an omission of type verification. Rather, by isolating type-related statements, we can safely and cleanly decouple the structural type constraints from the core deductive logic, making the natural language step significantly easier to evaluate.
\\

\noindent
For a practical way to formalize the elaboration process, we define an \textbf{explanation} of the step $s_i$ as a triple of sets $(\Gamma_i, \Sigma_i, \mathcal{T}_i)$, which encompasses all the necessary statements required to logically derive the assertion $a_i$. Thus for any elaboration $s_i'$(in natural language), there exists a explanation $(\Gamma_i, \Sigma_i, \mathcal{T}_i)$ that \textbf{explain} $s_i'$. In this context, we say the triple \textbf{explain}s the elaboration if every critical mathematical assertion articulated within the natural language elaboration and the step is subsumed either directly by the sets of the triple or by the intermediate statements deduced during the derivation process. Moreover, the derivation logically entailed by this triple must strictly adhere to the specific computational or deductive methodology dictated by the natural language text. $\Gamma_i$ contains finitely many statements (theorems, lemmas, definitions, and hypothesis) from $S$. The statements within $\Gamma_i$ may either be explicitly stated in $S$ or left implicit. $\Sigma_i$ comprises a finite set of statements drawn from external literature or established domain consensus, that is, outside of $S$. Here statements from $\Sigma_i$ are considered non-trivial. Within an explanation of $s_i'$, there may exist statements from $\Sigma_i$ that are implicitly assumed rather than explicitly articulated in $s_i'$. $\mathcal{T}_i$ comprises a finite set of what are often considered trivial statements (or rewrite rules) representing the background theory. It assists $\Gamma_i$ and $\Sigma_i$ in syntactically deriving the assertion $a_i$. Fundamentally, our framework aims to model any detailed elaboration of a proof step. Consequently, any mathematical statement invoked in such an expanded derivation must necessarily originate from this triple.
\\

\noindent
When we say a step $s_i$ is valid, we usually mean there exists an explanation $(\Gamma_i, \Sigma_i, \mathcal{T}_i)$ of $s_i$ such that:$$\Gamma_i, \Sigma_i, \mathcal{T}_i\vdash a_i$$ Here $\vdash$ denotes a rigorous informal entailment. Also, all $\Gamma_i, \Sigma_i, \mathcal{T}_i$ are valid. For $\Gamma_i$ to be valid, all the statements it contains should first be effective in the step $s_i$. Statements in $S$ has their scope of effectiveness. A context statement (an assumption or a declaration) has its scope: assumptions and variables are no longer valid from a certain point onwards. Consequently, $s_i$ may also fall outside the scope of other preceding mathematical statements, including lemmas and theorems. For example, if the proof contains a case analysis, then for every case the proof establishes a block with the corresponding assumption. The assumption's scope terminates once the discussion for the case is over. The theorems and lemmas proved in the case may also be invalid for the next case, if local assumptions and declarations are used. Second, any implicit statements within $\Gamma_i$ must be validly inferred from the context. The admissibility of such implicit elements is highly sensitive to both the active discourse and the established consensus of the mathematical community. Consequently, LLMs should be able to correctly identify valid implicit statements based on the context. Finally, since $\Gamma_i$ contains deductions and theorems from the context, the statements of $\Gamma_i$ should all be valid. For $\Sigma_i$ to be valid, all statements within it must be drawn from universally accepted mathematical facts or sound external literature.
\\

\noindent
We now explain the validity of $\mathcal{T}_i$. Since $\mathcal{T}_i$ captures the routine background steps skipped by human writers, it provides a practical interface between informal mathematical text and formal axioms. There is an immediate problem of what is considered as a nontrivial statement and has to be written explicitly. Since the explanation here actually corresponds to expanded derivations of $s_i$, the threshold for non-triviality can be carefully calibrated. Also, for $\mathcal{T}_i$ to be valid, all the statements it contains should be logically sound. Throughout this framework, we operate under the standard assumption of soundness for the underlying mathematical foundation. Let us now consider a simple example of a proof and two possible explanation of different steps. Note that by admitting the validity of the local context in eliminating ambiguity, we can omit type declarations for the assertions and the other statements.

\begin{tcolorbox}[
    colback=gray!5, 
    colframe=gray!50!black, 
    title=\textbf{Case Study: Deconstructing the Proof of $\sqrt{2} \notin \mathbb{Q}$},
    fonttitle=\bfseries,
    arc=2mm, 
    breakable
]
\textbf{Background Proof Context:} 

\textbf{Theorem:} $\sqrt{2}$ is an irrational number.

\noindent \textbf{Step 1 ($s_1$):}  \textit{Assume for the sake of contradiction that $\sqrt{2}$ is a rational number.}

\noindent \textbf{Step 2 ($s_2$):}  \textit{By the definition of rational numbers, there exist integers $a$ and $b$ (where $b \neq 0$) such that $\sqrt{2} = \frac{a}{b}$.}

\noindent \textbf{Step 3 ($s_3$):}  \textit{We can further assume that $a$ and $b$ are coprime (i.e., the fraction $\frac{a}{b}$ is in its simplest form, sharing no common factors other than 1).}

\noindent \textbf{Step 4 ($s_4$):}  \textit{Squaring both sides of the equation gives $2 = \frac{a^2}{b^2}$.}

\noindent \textbf{Step 5 ($s_5$):}  \textit{Multiplying both sides by $b^2$ yields $2b^2 = a^2$.}

\noindent \textbf{Step 6 ($s_6$):}  \textit{Since $b$ is an integer, $b^2$ is an integer, making $2b^2$ an even number. Therefore, $a^2$ must be an even number.}

\noindent \textbf{Step 7 ($s_7$):}  \textit{If the square of an integer is even, then the integer itself must be even. Thus, $a$ is an even integer.}

\noindent \textbf{Step 8 ($s_8$):}  \textit{Since $a$ is even, there exists an integer $k$ such that $a = 2k$.}

\noindent \textbf{Step 9 ($s_9$):}  \textit{Substituting $a = 2k$ into the equation from Step 5 ($2b^2 = a^2$) gives $2b^2 = (2k)^2 = 4k^2$.}

\noindent \textbf{Step 10 ($s_{10}$):} \textit{Dividing both sides by 2 yields $b^2 = 2k^2$.}

\noindent \textbf{Step 11 ($s_{11}$):} \textit{Following the same logic as before, this implies that $b^2$ is an even number.}

\noindent \textbf{Step 12 ($s_{12}$):} \textit{Consequently, $b$ must also be an even number.}

\noindent \textbf{Step 13 ($s_{13}$):} \textit{Since both $a$ and $b$ are even numbers, they share a common factor of at least 2.}

\noindent \textbf{Step 14 ($s_{14}$):} \textit{This directly contradicts our assumption in Step 3 that $a$ and $b$ are coprime.}

\noindent \textbf{Step 15 ($s_{15}$):} \textit{Therefore, our initial assumption in Step 1 must be false, concluding that $\sqrt{2}$ is irrational.}
\vspace{2mm}
\hrule
\vspace{2mm}

\textbf{Step 7 ($s_7$):}

\begin{itemize}
    \setlength{\itemsep}{1pt}
    \item \textbf{Assertion ($a_7$):} $a$ is an even integer.
    \item \textbf{Local Context ($\Gamma_7$):} $\{a^2 \text{ is even}\} \cup \{a \in \mathbb{Z}\}$ \hfill \textit{(Note: $a \in \mathbb{Z}$ is an implicit statement)}
    \item \textbf{Global Knowledge ($\Sigma_7$):} $\{\forall n \in \mathbb{Z}, n^2 \text{ is even} \implies n \text{ is even}\}$
    \item \textbf{Background Theory ($\mathcal{T}_7$):} $\emptyset$ \hfill \textit{(Direct application of Modus Ponens)}
\end{itemize}

\vspace{2mm}
\hrule
\vspace{2mm}

\textbf{Step 9 ($s_9$):} 

\begin{itemize}
    \setlength{\itemsep}{1pt}
    \item \textbf{Assertion ($a_9$):} $2b^2 = (2k)^2 =4k^2$.
    \item \textbf{Local Context ($\Gamma_9$):} $\{a = 2k, \quad 2b^2 = a^2\}$
    \item \textbf{Global Knowledge ($\Sigma_9$):} $\emptyset$ \hfill \textit{(No external lemmas invoked)}
    \item \textbf{Background Theory ($\mathcal{T}_9$):} Equivalence properties of equality (e.g., symmetry, transitivity, and substitution/congruence) combined with fundamental properties of integer arithmetic (e.g., commutativity, associativity, and evaluation of ground constants such as $2\times 2=4$).
\end{itemize}
\end{tcolorbox}

\noindent
We now analyze the benefits of not restricting $\mathcal{T}_i$ to rigid, hardcoded rules, primarily regarding its interface with foundational axioms. Although we do not specify a definitive axiomatic system, our framework inherently assumes universally permitted operations like Modus Ponens. This universal compatibility affords us considerable flexibility in formally specifying $\mathcal{T}_i$. Furthermore, the flexibility in specifying $\mathcal{T}_i$ drastically reduces the deductive overhead within $\vdash$ and minimizes the total number of explicit statements required. For example, in formal libraries like Lean’s \texttt{Mathlib}, performing basic integer arithmetic typically necessitates instantiating deep algebraic hierarchies, such as commutative rings. In contrast, allowing all kinds of sound statements bypasses this formal complexity, enabling the system to directly reason with sound statements over $\mathbb{Z}$. Furthermore, since the operations within $\mathcal{T}_i$ are inherently routine, we can leverage this flexibility to formulate statements directly about the concrete objects, bypassing the need for explicit variable substitution. This also mirrors human conventions in mathematical discourse: when executing elementary transformations, we naturally manipulate the expressions directly. Consequently, the elements within $\mathcal{T}_9$ are concrete statements like $(2k)^2 = 2k \cdot 2k$, rather than abstract, uninstantiated schemas such as $x^2 = x \cdot x$.
\\

\noindent
Note that the explicit type declaration $a^2 \in \mathbb{Z}$ is omitted from $\Gamma_7$ for the same reason we bypass redundant variable bindings. By the ambient context, we establishes $a \in \mathbb{Z}$. And since $\{\forall n \in \mathbb{Z}, n^2 \text{ is even} \implies n \text{ is even}\}$ is an unambiguous well-established statement under the current context, we don't need to write $a^2 \in \mathbb{Z}$ here. Consequently, an explanation $(\Gamma_i, \Sigma_i, \mathcal{T}_i)$ should not be viewed as the complete set of all contextually valid truths, but exclusively as the minimal required set of statements necessary to validate the natural language proof step rigorously.
\\

\noindent
In the example above, we directly structured an elaboration using the $(\Gamma_i, \Sigma_i, \mathcal{T}_i)$ format. In what follows, we demonstrate that given an elaboration $s_i'$ of a proof step, we can extract its constituent statements and systematically categorize them into this triple framework. Then, find an explanation that explains $s_i'$, though the explanation may not be valid. Consider the following example from Firstproof generated by Aletheia.
\\

\noindent

\begin{tcolorbox}[
    enhanced,
    breakable, 
    colback=gray!2, 
    colframe=black!70, 
    boxrule=0.6pt, 
    arc=0pt, 
    outer arc=0pt,
    left=2.5mm, right=2.5mm, top=2mm, bottom=2mm, 
    title=\textbf{Case Study: Extraction Of Statements}, 
    coltitle=black, 
    colbacktitle=gray!15, 
    titlerule=0.6pt, 
    fontupper=\small
]

\textbf{Problem and Proof Sketch} \\
\textit{Problem:} Suppose that $\Gamma$ is a uniform lattice in a real semi-simple group, and that $\Gamma$ contains some 2-torsion. Is it possible for $\Gamma$ to be the fundamental group of a compact manifold without boundary whose universal cover is acyclic over the rational numbers $\mathbb{Q}$? \\
\textit{Proof:} The generated proof constructs a logical contradiction centered around the compactly supported Lefschetz number, $L_c(\gamma, X)$. The model computes this topological invariant through two divergent pathways. Algebraically, it correctly applies Poincaré duality alongside the $\mathbb{Q}$-acyclicity of $X$ to deduce that $H_c^*(X; \mathbb{Q})$ is concentrated exclusively in the top degree, cleanly evaluating to $L_c(\gamma, X) = \pm 1 \neq 0$. Topologically, the model evaluates the 2-sheeted regular covering $X \to Y = X/\langle\gamma\rangle$ using the transfer map, arriving at the valid relation $\chi_c(Y) = \frac{1}{2}(\chi_c(X) + L_c(\gamma, X))$. However, to mathematically force a contradiction, the model introduces a critical hallucination: it erroneously asserts that the compactly supported Euler characteristic is strictly multiplicative (i.e., $\chi_c(X) = 2\chi_c(Y)$) for infinite cell complexes. By substituting this flawed generalization into the transfer equation, the generated proof strictly forces $L_c(\gamma, X) = 0$, resulting in the false algebraic contradiction $0 = \pm 1$.

\vspace{2mm}\noindent\textcolor{gray}{\rule{\linewidth}{0.5pt}}\vspace{2mm}

\textbf{Step $s_i$ (Flawed)} \\
\textit{"However, for any finite $k$-sheeted unbranched covering of locally compact ENRs with finite Betti numbers, the compactly supported Euler characteristic is strictly multiplicative. Thus, $\chi_{c}(X)=2\chi_{c}(Y)$."}

\vspace{2mm}\noindent\textcolor{gray}{\rule{\linewidth}{0.5pt}}\vspace{2mm}
 \textbf{Explanation generated by LLM}\\
    \sethlcolor{green!20}
    The author introduces a general rule regarding covering spaces. They assert:
    \begin{quote}
        \hl{\textit{``for any finite $k$-sheeted unbranched covering... the compactly supported Euler characteristic is strictly multiplicative.''}}
    \end{quote}

    \sethlcolor{yellow!20}
    \hl{Expressed as a mathematical formula, the author is asserting that for any total space $E$ and base space $B$ connected by a $k$-sheeted covering map, the relationship is:}
    \begin{center}
    \colorbox{yellow!30}{$ \chi_c(E) = k \cdot \chi_c(B) $}
    \end{center}

    To use this general formula, the author must establish the specific values for $E$, $B$, and $k$ based on the geometry of the spaces in the proof.
    \begin{itemize}
        \item \textbf{The Total Space ($E$):} The author is using $X$ (the universal cover of the manifold $M$) as the total space. 
        \item \textbf{The Base Space ($B$):} The author defines the base space $Y$ as the quotient $X/\langle\gamma\rangle$. 
        \sethlcolor{blue!20}
        \item \textbf{The Sheet Count ($k$):} The author establishes that $\gamma$ is a torsion element of order 2 ($\gamma^2 = \text{id}$). Because $\gamma$ generates a strictly free action of $\mathbb{Z}/2\mathbb{Z}$ on $X$, \hl{the projection map from $X$ down to the quotient $Y$ is a regular covering map.} Because the group acting on it has exactly 2 elements, \hl{the number of ``sheets'' (the size of the fiber over any point in $Y$) is exactly 2. Therefore, $k = 2$.}
    \end{itemize}
    
    The author takes the specific parameters established in Step 2:
    \begin{itemize}
        \item $E \to X$
        \item $B \to Y$
        \item $k \to 2$
    \end{itemize}
    
    And substitutes them directly into the general formula from Step 1 ($\chi_c(E) = k \cdot \chi_c(B)$). 
    
    This direct application of the parameters to the multiplicative rule yields the final assertion:
    \[ \chi_c(X) = 2\chi_c(Y) \]
    
    That is the entirety of the deduction for that specific step. The author simply proves the covering is 2-sheeted and plugs $k=2$ into their (flawed) multiplicative formula.

\vspace{2mm}\noindent\textcolor{gray}{\rule{\linewidth}{0.5pt}}\vspace{2mm}

\textbf{Statement Extraction $(\Gamma_i, \Sigma_i, \mathcal{T}_i)$} \\
\textbf{Assertion ($a_i$):} $\chi_{c}(X)=2\chi_{c}(Y)$.\\
\textbf{Local Context ($\Gamma_i$):} \\
$\gamma_1$ : The projection $X\rightarrow Y$ is a 2-sheeted regular covering map, generated by a strictly free group action.\\
$\gamma_2$ : $H_c^*(X; \mathbb{Q})$ and $H_c^*(Y; \mathbb{Q})$ are finite-dimensional.\\
$\gamma_3$ : $X$ is an $n$-dimensional topological manifold. $Y$ is a topological manifold.\\
All of them are explicit.\\
\textbf{Global Knowledge ($\Sigma_i$):} \\
$\sigma_1$ : \textit{[Flawed lemma]} For any finite $k$-sheeted unbranched covering of locally compact ENRs with finite Betti numbers, the compactly supported Euler characteristic is strictly multiplicative. \\
$\sigma_2$ : Every finite-dimensional topological manifold is a locally compact ENR.\\
$\sigma_3$ : A topological manifold has finite compactly supported Betti numbers if its compactly supported rational cohomology is finite-dimensional.\\
$\sigma_4$ : If $X \to Y$ is a covering map between manifolds, then $\dim(Y) = \dim(X)$.\\
\textbf{Background Theory ($\mathcal{T}_i$):} \\
$t_1$ : The property of 'the Euler characteristic being strictly multiplicative' is defined by the following equation under algebraic rewrite rules: $\chi(\text{total space}) = k \cdot \chi(\text{base space})$, where $k$ is the number of sheets of the covering.\\
$t_2$ : A regular covering map generated by a strictly free group action is an unbranched covering.\\
$t_3$ : A k-sheeted covering is finite.

\vspace{2mm}\noindent\textcolor{gray}{\rule{\linewidth}{0.5pt}}\vspace{2mm}

\textbf{Deduction $\vdash$}\\
Since $\gamma_1$ establishes a covering map $X \to Y$ and $\gamma_3$ identifies both as manifolds, $\sigma_4$ confirms they share the same dimension $n$, which by $\sigma_2$ ensures both are locally compact ENRs. Concurrently, because $\gamma_2$ confirms their compactly supported rational cohomology is finite-dimensional, $\sigma_3$ guarantees both possess finite compactly supported Betti numbers. Given that $\gamma_1$ specifies the action is strictly free, $t_2$ classifies it as unbranched, and since it is a 2-sheeted covering, $t_3$ confirms it is finite $k$-sheeted ($k=2$). With all prerequisites of the flawed lemma $\sigma_1$ met (a finite $k$-sheeted unbranched covering of locally compact ENRs with finite Betti numbers), we apply $t_1$ with $k=2$ to yield the final assertion $a_i$: $\chi_{c}(X) = 2\chi_{c}(Y)$.

\end{tcolorbox}

\noindent
The green highlighted part corresponds to $\sigma_1$. The yellow highlighted statement corresponds to $t_1$. The blue highlighted statement corresponds to $\gamma_1$. Notice a few things here. First, we ignore the statements ``$\gamma$ generates a strictly free action of $\mathbb{Z}/2\mathbb{Z}$  on $X$,'' and ``the group acting on it has exactly 2 elements.'' But we still consider our $(\Gamma_i, \Sigma_i, \mathcal{T}_i)$ explains the elaboration. By definition, the explanation should contain every critical mathematical assertion articulated within the natural language text. Thus, by considering the above two statements to be not critical (which is reasonable), we can simplify our $(\Gamma_i, \Sigma_i, \mathcal{T}_i)$. Given the inherent linguistic variations and rhetorical scaffolding present in human-authored proofs, the extraction of critical statements from natural language inherently demands a degree of flexibility. Second, we did not take arguments like $E$ is $X$ into account. This is also for simplification. By the same manner of the first example, we omit the detail of Modus Ponen. The generated elaboration doesn't verify all required conditions (for example, finite Betti number) for the flawed theorem, thus we can fill them up in the explanation. Though the deduction process $\vdash$ of the explanation is sound, the statements it contains are not valid, making the explanation itself not valid. For certain steps, the logical deduction is sufficiently explicit that it requires almost no hidden calculations, rendering the corresponding elaboration essentially deterministic in practice.

In summary, this work seeks to analyze the rigor of natural language mathematical proofs. We do not attempt to enforce the absolute strictness required by systems like the Mathematical Vernacular (MV) or formal proof assistants. Empirically, however, we demonstrate that subtle logical errors can be successfully localized simply by writing out the proof in granular detail. The critical factor in this approach is maintaining strict faithfulness to the author's original intent. Ultimately, our theoretical framework serves primarily as philosophical guidance for the agent's reasoning, rather than a rigid requirement to perfectly instantiate every formal explanation. As our experiments confirm, the mere act of forcing the agent to elaborate in detail intrinsically elevates its verification performance.

\section{Case Study: The Boundary of Implicit Conventions (Aletheia 7A \& 7B)}

To gain deeper insights into the agent's cognitive trajectory, we present a detailed comparative case study on the proofs \texttt{Aletheia 7A} (invalid) and \texttt{Aletheia 7B} (valid). 

The original text of \texttt{7A} actually consists of two separate proofs: the first is an invalid pseudoproof containing a highly concealed, fatal error concerning ``multiplicity'', while the second fundamentally relies on the first. For GPT-5.4, formulating a precise disproof for this specific error is difficult, as explicit counterexamples are scarcely available via web retrieval and fall outside the model's internal parametric knowledge. Furthermore, within the specific scope of FirstProof Problem 7, it is an accepted academic convention to assume that semisimple Lie groups are linear—an assumption that does not strictly hold in general mathematical contexts. Consequently, an unguided agent is highly susceptible to being distracted by this linearity assumption (generating a false flaw witness) or by downstream errors in the second proof.

To stress-test our agent's ability to identify the true structural error, we manually truncated \texttt{7A} to include only the first proof and explicitly appended the linearity condition. Across 10 independent runs, the agent yielded one false negative (marking the flawed step as \texttt{VALID} during initialization rather checking it carefully) and successfully rejects the proof 9 times, marking the specific step \texttt{OPEN} or \texttt{FLAWED}. However, a closer inspection reveals a critical logical subtlety: the agent justified its \texttt{FLAWED} verdict by pointing out that the multiplicity claim explicitly contradicted the local context. While this contradiction genuinely exists, penalizing the step strictly on these grounds is mathematically unreasonable here, as the step resides within a proof by contradiction (\textit{reductio ad absurdum})—where deriving absurdities is the expected trajectory. This highlights the profound difficulty in distinguishing between an invalid inference rule and a valid derivation of a contradiction.

Conversely, our experiments on the fundamentally valid proof \texttt{Aletheia 7B} perfectly illuminate the agent's underlying verification power and its dynamic self-correcting capabilities. Though the agent fail to validate the proof in the 3 runs, when we evaluated \texttt{7B} by explicitly providing the linearity assumption and the reference solution sketch, the agent did not simply rubber-stamp the text. Instead, it exhibited a rigorous, non-linear cognitive trajectory. 

During the verification process, the agent initially proposed two potential flaw witnesses that threatened to invalidate the proof. In one of them, the agent's web retrieval module returned an alternative formulation of a specific series expansion. Detecting a difference from the retrieved literature, the agent instinctively marked the author's step as \texttt{FLAWED}. The action is then rejected by our \textit{Flaw Confirmation} mechanism. Engaged in adversarial self-reflection, the agent re-evaluated the discrepancy and recognized that the mere existence of a different formalization online does not mathematically invalidate the author's specific local derivation. Consequently, the agent correctly dismissed its own false alarm, averting a premature false positive. 

After overriding these superficial discrepancies and conducting further targeted retrieval, the agent meticulously verified the highly non-trivial claims regarding KO-theory (real topological K-theory) and successfully closed the proof. 

This juxtaposition offers a more grounded observation: when explicit premises are provided, the agent is capable of processing complex mathematical concepts like K-theory. More importantly, this iterative pattern—searching for external literature, raising potential issues, and self-correcting—demonstrates a practical pathway for scaling test-time compute in mathematical verification. By allocating more inference steps to information gathering and adversarial reflection, the system can progressively resolve textual ambiguities. However, the agent's frequent stalling on implicit human conventions serves as a humbling reminder. It indicates that while deductive logic and external retrieval are necessary, they are not yet sufficient to fully bridge the gap between strict formalization and the nuanced, context-dependent nature of human mathematical communication.

\section{Prompts for the Sentence-by-Sentence Ablation Baseline}
\label{appendix:ablation_prompts}

In this section, we present the prompts used for the \textbf{Sentence-by-Sentence} ablation baseline discussed in Section 4.1. 

Unlike our main Constructive Verification Agent, which strictly enforces the $(\Gamma_i, \Sigma_i, \mathcal{T}_i)$ tripartite structure and maintains an explicit external theorem ledger, this ablation baseline is designed to test the effect of mere textual granularity. Specifically, it instructs the agent to break the proof down into sentence-level steps but only requests a ``plain naive analysis in prose'' without rigid dependency tracking. This isolates the performance gains achieved by our structured deductive constraints from the gains naturally occurring when LLMs process text at a granular level. To ensure a fair and controlled ablation, the overall execution pipeline—including the reflection and flaw confirmation prompts—remains identical to our primary version.

\vspace{1em}

\begin{tcolorbox}[promptbox, title=Ablation Prompt 1: Initialization]
\ttfamily\small
Run the verifier in this folder. Read only \texttt{README.md} and \texttt{source/input.txt}. Split the proof into small steps. Create one \texttt{.txt} file in \texttt{steps/} for each step. Every original source sentence with mathematical content must appear verbatim in at least one step file. Initialize or refresh \texttt{steps/}, \texttt{stepsprocess.txt}, and \texttt{answer.txt}.

On the first pass, do not read \texttt{source/theory.txt}, \texttt{source/example.txt}, \texttt{source/standards.txt}, or \texttt{reference/*}.

Work step by step. In each step file, keep the original text and then write a plain naive analysis in prose of whether the step closes, what is missing, or whether there is an explicit flaw witness. Be conservative: if the step is not convincingly closed, keep it \texttt{OPEN}. Use \texttt{FLAWED} only with an explicit flaw witness.

Maintain only \texttt{steps/}, \texttt{stepsprocess.txt}, and \texttt{answer.txt}. Stop only when every step has been marked \texttt{VERIFIED}, \texttt{OPEN}, or \texttt{FLAWED}.
\end{tcolorbox}

\vspace{1em}

\begin{tcolorbox}[promptbox, title=Ablation Prompt 2: Resume]
\ttfamily\small
Resume the verifier in this folder. Read \texttt{README.md}, \texttt{stepsprocess.txt}, \texttt{answer.txt}, and all unresolved files in \texttt{steps/}. If a step is still too coarse, split it smaller. Every original source sentence with mathematical content must appear verbatim in at least one step file.

Continue step by step. If a step has remained \texttt{OPEN} since the first pass, consider whether there is an explicit flaw witness and it should be \texttt{FLAWED}. For a hard \texttt{OPEN} step, you may consult \texttt{reference/*} or search online for relevant details. Before changing any unresolved step to \texttt{VERIFIED}, recheck that the step really closes.

Default negative verdict is \texttt{OPEN}. Use \texttt{FLAWED} only with an explicit flaw witness. Maintain only \texttt{steps/}, \texttt{stepsprocess.txt}, and \texttt{answer.txt}.
\end{tcolorbox}

\section{Implementation Details and Hyperparameters}
\label{appendix:hyperparameters}

To ensure full reproducibility of our empirical results, we detail the exact execution environment and API parameter configurations utilized for our framework. 

\paragraph{Execution Environment.} All experimental evaluations were conducted within the \texttt{codex} framework (version 0.124.0). We strictly utilized the framework's default configuration profile for environmental setup. Beyond this basic scaffolding, all complex agentic behaviors, formatting rules, and state-machine constraints were injected directly into the GPT-5.4-xhigh engine via natural language system prompts. This design avoids overfitting to specialized framework configurations and explicitly demonstrates the out-of-the-box generalizability of our verification approach.

\paragraph{LLM Generation Hyperparameters.} To maintain a rigorously fair comparison, identical generation parameters were applied to the GPT-5.4-xhigh API calls across all baselines and our agentic pipeline. We deliberately operated at a standard temperature of 1.0 to simulate natural generation variance. The fact that our agent yields deterministic, monotonic convergence under these high-entropy sampling conditions further corroborates the robustness of our prompt-driven constraints. The complete parameter suite governing the LLM API requests is summarized in Table \ref{tab:hyperparameters}. 

\begin{table}[h]
\centering
\caption{Complete hyperparameter configurations applied directly to the GPT-5.4-xhigh API endpoints across all verification pipelines.}
\label{tab:hyperparameters}
\small
\setlength{\tabcolsep}{4pt}
\renewcommand{\arraystretch}{1.2}
\begin{tabular}{lc}
\toprule
\textbf{Hyperparameter} & \textbf{Value} \\
\midrule
Temperature & 1.0 \\
Top-$p$ & 1.0 \\
Top-$k$ & 0 \\
Min-$p$ & 0.0 \\
Top-$a$ & 0.0 \\
Frequency Penalty & 0.0 \\
Presence Penalty & 0.0 \\
Repetition Penalty & 1.0 \\
\bottomrule
\end{tabular}
\end{table}

\section{A Worked Trace of a Successful Verification Run for Problem 7b}
\label{app:7b-trace}

This appendix records a representative successful verification trace for Problem~7b, based on the segment of the execution log that culminates in a fully verified proof state. The point of this example is not that the agent generated a correct proof in one shot. Rather, the log shows a staged verification process in which the agent repeatedly refines the proof graph, isolates unresolved dependencies, revises an incorrect intermediate diagnosis, and only then closes the remaining steps.

The successful portion of the run is the segment in which the global status eventually changes to \texttt{Overall verdict: VERIFIED}. Within that segment, the agent's behavior is best understood as a sequence of state transitions over a ledger of proof obligations, rather than as a single forward derivation.

\subsection{Initial Reduction of the Open Proof State}

At the beginning of the successful segment, the proof is not globally closed. The agent first strengthens the verification of the real assembly injectivity step, recorded as \texttt{step\_15a\_real\_assembly\_injectivity}. The key effect of this move is structural: it eliminates one major upstream dependency and reduces the remaining unresolved part of the proof to the signature-package portion.

The agent then observes that the existing step \texttt{step\_15b1\_signature\_equals\_assembly} is too coarse. In its own diagnosis, this step mixes two logically different bridges:
\begin{enumerate}[label=(\roman*)]
    \item the \textit{assembly/index bridge}, namely that the analytic symmetric signature is the assembly image of a signature class; and
    \item the \textit{characteristic-class bridge}, namely that this signature class corresponds rationally to the Poincar\'{e} dual of the Hirzebruch $L$-class.
\end{enumerate}

Instead of keeping this as one unresolved block, the agent explicitly resplits it into smaller units:
\begin{itemize}
    \item \texttt{step\_15b1a\_signature\_class\_to\_assembly}
    \item \texttt{step\_15b1b\_signature\_class\_to\_lclass}
    \item \texttt{step\_15b1c\_signature\_equals\_assembly}
\end{itemize}

It also rewires the downstream chain so that each later step depends on the newly separated obligations rather than on one monolithic unresolved assertion:
\begin{center}
    \small\ttfamily
    step\_15b2\_m0\_pushforward\_formula $\rightarrow$ step\_15c\_lclass\_equality $\rightarrow$ \\
    step\_16\_degree\_pm1 $\rightarrow$ step\_17\_final\_contradiction
\end{center}

This resplitting is the first decisive move in the successful run. It turns a vague unresolved block into a small dependency graph with identifiable root causes.

\subsection{Compression to Two Genuine External Gaps}

After the split, the proof still contains several \texttt{OPEN} steps, but the agent recognizes that these are largely downstream consequences of only two genuine external gaps, recorded in the dependency ledger as \texttt{D6b1a} and \texttt{D6b1b}. Informally:
\begin{itemize}
    \item \textbf{\texttt{D6b1a}:} the real $KO$-theoretic assembly/index bridge.
    \item \textbf{\texttt{D6b1b}:} the real $KO$-homological $L$-class bridge.
\end{itemize}

This is an important feature of the trace. The raw number of open files is not itself the right complexity measure. Once the proof has been resplit correctly, the open graph becomes shallow: many open steps are merely inherited consequences of a very small number of unresolved inputs.

\subsection{An Incorrect Flaw Diagnosis and Its Retraction}

The run then exhibits a nontrivial self-correction. The agent temporarily marks \texttt{step\_15b1c\_signature\_equals\_assembly} as \texttt{FLAWED}. The reason given is a purported contradiction with a Rosenberg--Weinberger normalization theorem, which seemed to show that the exact real $KO$-theoretic formula written in the step could not hold with coefficient $1$ and the Hirzebruch $L$-class exactly as written.

This is not the final diagnosis. A separate follow-up phase is launched specifically to determine whether the recorded flaw witness really implies that the original step is false as used. This matters because the verification protocol distinguishes sharply between:
\begin{itemize}
    \item a statement that is \textit{genuinely false};
    \item a statement that may still be true, but whose current justification is \textit{incomplete or misaligned} with the cited source.
\end{itemize}

Upon re-reading the relevant source, the agent concludes that the previous witness targeted the signature-operator transformation $s_n$, whereas the proof step under inspection was using a symmetric-signature route. Therefore the witness did \textbf{not} actually refute the exact as-used inference. The result is a rollback from \texttt{FLAWED} to \texttt{OPEN}.

This retraction is one of the most informative parts of the trace. It shows that the agent is not merely searching for counterexamples, but is also checking whether a counterexample hits the correct formal target. In other words, the system distinguishes a failed proof route from a failed theorem statement.

\subsection{Closure of the Two Remaining Root Dependencies}

After retracting the incorrect flaw diagnosis, the agent returns to the two root gaps.

\paragraph{1. Closing \texttt{D6b1a}.} The agent verifies \texttt{step\_15b1a\_signature\_class\_to\_assembly}. The successful route here is to replace an indirect complex-$K$-theoretic justification with a more exact real $KO$-theoretic source: Rosenberg and Weinberger's Lipschitz-signature theorem is used to construct the real $KO$-homology signature class and to identify its assembly image with the symmetric signature. Once this step is marked \texttt{VERIFIED}, the remaining active dependency set shrinks from \{\texttt{D6b1a}, \texttt{D6b1b}\} to just \{\texttt{D6b1b}\}.

\paragraph{2. Closing \texttt{D6b1b}.} The agent closes this gap by rewriting \texttt{step\_15b1b\_\allowbreak signature\_class\_to\_lclass}. Here the successful move is not simply to cite a missing theorem verbatim. Instead, the agent builds a local bridge:
\begin{enumerate}[label=(\arabic*)]
    \item identify the real signature class in $KO$-homology;
    \item complexify it to the standard complex signature class;
    \item invoke the known complex $K$-homological $L$-class formula;
    \item use rational real/complex comparison to recover the required real $KO$-homology coordinates.
\end{enumerate}

This is the step at which the open graph finally collapses. Once \texttt{step\_15b1b\_signature\_class\_to\_lclass} is verified, the recombination step \texttt{step\_15b1c\_signature\_equals\_assembly} also becomes verifiable, since its two incoming components are now both closed.

\subsection{Propagation to the Final Contradiction}

After the two root dependencies are closed, the rest of the proof is discharged by dependency propagation rather than by new conceptual discoveries.
\begin{itemize}
    \item \texttt{step\_15b1c\_...} is verified by recombining the now-verified assembly/index bridge with the $L$-class bridge.
    \item \texttt{step\_15b2\_...} is verified by specializing the base formula to $M_0$ and $X_0$.
    \item \texttt{step\_15c\_...} is verified by using injectivity of the real assembly map to pull back the signature equality.
    \item \texttt{step\_16\_...} is verified by reading off the top-degree consequence, namely that $D=\pm 1$.
    \item \texttt{step\_17\_...} is finally verified by combining this conclusion with the already verified evenness statement for $D$.
\end{itemize}
\textit{(Note: Long file names are abbreviated here for narrative flow, while remaining rigorously tracked in the system ledgers.)}

At that point, the global ledger is updated to \texttt{remaining open steps: none} and \texttt{Overall verdict: VERIFIED}. The run concludes with a consistency pass over the step files, the dependency ledger, the process summary, and the final answer file to ensure that no stale \texttt{OPEN} or \texttt{FLAWED} status remains.

\subsection{Interpretation}

This trace is useful because it illustrates what a successful agentic proof-verification run actually looks like in practice. It is not a single-pass proof synthesis. Instead, it has four characteristic ingredients:
\begin{enumerate}[label=\textbf{Phase \arabic*:}, leftmargin=*]
    \item \textbf{State Decomposition:} A coarse unresolved claim is split into minimal logical units.
    \item \textbf{Dependency Localization:} Many apparent open steps are reduced to a very small set of root gaps.
    \item \textbf{Error Correction:} An intermediate \texttt{FLAWED} diagnosis is explicitly re-examined and retracted when the witness is found to miss the exact target statement.
    \item \textbf{Closure by Propagation:} Once the root gaps are closed, the remaining steps verify in sequence through the dependency graph.
\end{enumerate}

For this reason, the \texttt{7B} example should not be described as a case where the agent simply ``found the proof.'' A more accurate description is that it executed a ledger-guided verification procedure: it refined the proof graph, isolated the true bottlenecks, corrected a mistaken local refutation, and only then propagated the repaired information to a fully verified final contradiction.

\end{document}